# Time, Chance, and Action

**Peter Haddawy** *
Department of Computer Science, University of Illinois
405 North Mathews Avenue, Urbana, Illinois 61801
email: haddawy@cs.uiuc.edu

## Abstract

To operate intelligently in the world, an agent must reason about the consequences of its actions. The consequences of an action are a function of both the state of the world and of the action itself. In realistic domains we cannot expect to have complete information, so a representation for reasoning about actions must be able to express uncertainty concerning the state of the world and the effects of actions and other events. This paper presents a future-branching temporal probability logic for reasoning about actions. The logic can represent the probability that facts hold and events occur at various times. It can represent the probability that actions and other events affect the future. It can represent concurrent actions and conditions that hold or change during execution of an action. The model of probability relates probabilities over time. The logical language integrates both modal and probabilistic constructs and can thus represent and distinguish between possibility, probability, and truth. Several examples illustrating the use of the logic are given.

## 1 Introduction

To choose intelligent courses of action, an agent must reason about the state of the world and the way in which actions affect the world. In realistic domains we cannot expect to have complete information, so a representation for reasoning about actions must be able to express uncertainty concerning the state of the world and the effects of actions and other events. For example, the statement "If I were to smoke, I would contract lung cancer some years down the road" can at best be said to hold with high likelihood. There are uncertain environmental factors that can influence my chance of contracting cancer as well as uncertainty in the effects of smoking.

In order to reason about the effects of actions, it is necessary to be able to reason about time. Facts tend to be true for periods of time and events occur at particular times. Actions comprising a plan may occur sequentially or concurrently. Actions and other events affect the future, but not the past. Chance evolves with time: the chance of rain tomorrow may not be the same now as it will be tonight. Ambiguities in the world are resolved with time: before a fair coin is flipped the chance of heads is 50% but after it is flipped it either certainly landed heads or it certainly did not.

We present a propositional temporal probability logic that can represent all of these aspects of time and chance. The logical language integrates both modal and probabilistic constructs and can thus represent and distinguish between possibility, probability, and truth. For example, the language allows us to write sentences that
1) describe concurrent actions:

- It is not possible for me to raise and lower my arm at the same time.

2) describe conditions during an action that influence the probabilistic effects of the action:

- If the oven temperature is constant while I am baking the souffle, the souffle is likely to turn out right.

3) mix statements about probability and inevitability:

- There is a 50% chance that at noon the train crash will be inevitable.

4) distinguish between truth and probability:

- I won the lottery even though it was unlikely.

Numerous researchers have developed temporal logics for reasoning about plans and actions [McDermott, 1982; Allen, 1984; Haas, 1985; Pelavin and

*This research was supported by the author's Cognitive Science/Artificial Intelligence fellowship from the University of Illinois. The author would like to thank Alan Frisch, Steve Hanks, Joe Halpern, Mike Wellman, Rich Scherl, Sheldon Nicholl, and Carl Kadie for helpful comments on earlier drafts of this paper.

Allen, 1986; Shoham, 1987]. Others have developed logics of probability [Fagin and Halpern, 1988; Bacchus, 1988]. But no work has addressed all theses issues in a comprehensive logical framework. Such a framework is necessary for representing and reasoning about plans in uncertain domains.

The theory literature contains several examples of logics that can represent both time and probability [Lehmann and Shelah, 1982; Hart and Shair, 1984; Halpern and Tuttle, 1989]. The focus of these logics is on reasoning about probabilistic programs and distributed systems. The logics do not attempt to model aspects of causality or to distinguish between different types of temporal objects such as facts and events; hence, they are not suitable for reasoning about actions and plans.

The logic presented here extends the capabilities of current planning logics by providing a probabilistic treatment of causality, concurrent actions, and conditions that hold or change during execution of an action. The logic combines aspects of Pelavin's [1988] temporal planning logic, van Fraassen's [1980] models of objective chance, and Fagin, Halpern, and Megiddo's [1988] probability logic. This paper does not address some of the traditional planning issues such as qualification and persistence. These issues are beyond the scope of the present work. What the developed logic does is provide a framework within which to explore such issues for domains in which uncertainty is a factor. Furthermore, this paper does not discuss complete axiomatizations of the logic. The logic presented is intended to be used as a tool in the representation of planning problems and the design and analysis of planning algorithms. The author does not believe that axiomatizing the logic and feeding the axioms to a theorem prover is a reasonable way to solve planning problems. Rather, we should strive to design special purpose planning algorithms that are faithful to the semantics of the logic. This logic presented will enable us to do this for a new class of interesting and difficult planning problems and to prove the correctness of these algorithms.

## 2 The Ontology

Time is modeled as a collection of world-histories, each representing one possible chronology or history of events throughout time. At any given point in time, some of the world-histories will naturally share a common past up to that time. Thus the world-histories can be formed into a tree structure that branches into the future. Note that there is no special status given to the time "now", so the temporal tree branches into the future relative to each point in time. The present work is only concerned with future-branching time because actions and other events can only affect the state of the world at times after their occurrence. Relative to any point in time, the past is determined; only the future contains open possibilities. We talk about specific times by reference to a global time line, a linearly-ordered set of points. The time line is necessary in order to compare facts and events at the same time in different branches of the time tree.

In the world, facts tend to hold and events tend to occur over intervals of time. Each fact is associated with the set of temporal intervals over which it holds. If a fact holds over an interval then it holds over all subintervals of that interval. So, for example, if my car is red over a period of time, then it is red throughout that time. Events are somewhat more complex than facts. First, one must distinguish between *event types* and *event tokens*. An event type is a general class of events and an event token is a specific instance of an event type. For example, picking up a book is an event type, and picking up the blue book on my desk at 9:00am is an event token of that type. The interval over which an event token occurs is the smallest interval over which it occurs from beginning to end. If I pick up a book during a time period, there is no smaller period of time during which the event of my picking up the book can be said to have occurred. On the other hand, numerous tokens of a given event type may occur during a particular interval. For example, I can pick up one book with my right hand and one with my left hand concurrently. So if a token of an event type occurs over an interval, it is possible for another token of that type to occur over a subinterval, but it is not necessary as it is in the case of facts. This paper will deal with event types and for brevity will simply refer to them as events.

The fact/event dichotomy just described is a simplification of the true situation. As Shoham [1987] has shown, there are many different types of facts and events, characterized by their temporal properties. Although Shoham's refined categories of fact types constitute a more useful and accurate picture of the world than the simple fact/event dichotomy, the fact/event categorization will be used for simplicity of exposition. Extending the work to encompass Shoham's categories is completely straightforward.

Uncertainty is represented by defining probabilities over the tree of possible futures. Because the present work is concerned with representing probabilistic effects, one talks about probability at a given point in time and probability is represented in such a way that the probability of the past, relative to that point, is either zero or one. In this way, actions and other events can only affect the probabilities of future facts and events. This type of probability is objective, as opposed to subjective probability, with regard to which the past can be uncertain. For example, subjectively I may be uncertain as to whether the train left on time or not, yet objectively it either certainly left or it certainly did not, and furthermore, there is nothing I can do now to change that. The other property imposed on objective probability is that the probability be completely determined by the history up to the current time. So probability is purely a function of the state of the world. This is in contrast to subjective probability where probability is a function not only of the state of the world but also of the epistemic state of an agent.

## 3 Syntax

To refer to facts and events occurring in time the language contains two predicates. $HOLDS$ associates a fact with the interval of time over which it is true and $OCC$ associates an event with the interval over which it occurs. The language contains three modal operators to talk about inevitability, possibility, and chance. The $\Box$ operator designates inevitability and we write $\Box_t(\phi)$ to indicate that $\phi$ is necessarily true at time $t$. Possibility $\Diamond$ is defined in terms of inevitability as $\Diamond_t(\phi) \equiv \neg\Box_t(\neg\phi)$. We write $\Diamond_t(\phi)$ to say that $\phi$ is possibly true at time $t$. The chance operator $P$ designates the probability of a sentence at a given time. We write $P_t(\phi) \geq \alpha$ to say that the probability of $\phi$ at time $t$ is at least $\alpha$. The sentence form $P_t(\phi) \leq \alpha$ is used as a shorthand for $P_t(\neg\phi) \geq 1-\alpha$, and similarly $=$, $>$, and $<$ are used. Following the syntax of Fagin, Halpern, and Megiddo's [1988] probability logic of polynomial weight formulas, polynomial combinations of probability operators will be allowed. Thus the language can express things like "$\phi$ is at least twice as likely as $\psi$": $P_t(\phi) \geq 2P_t(\psi)$. This is particularly useful for writing sentences about conditional probability. The probability of $\phi$ given $\psi$ is defined as

$$prob(\phi|\psi) = prob(\phi \wedge \psi)/prob(\psi).$$

If the probability of the conditioning sentence $\psi$ is zero, then the conditional probability is undefined. In this case, a conditional probability sentence like $prob(\phi|\psi) = \alpha$ can be assigned neither the value true nor the value false. Rather than introducing a new conditional probability operator and dealing with this truth assignment problem, sentences about conditional probability can simply be written in the form

$$P_t(\phi \wedge \psi) \geq \alpha P_t(\psi).$$

Note that the standard conditional probability notation will be used to syntactically denote a sentence of the above form:

$$P_t(\phi|\psi) \geq \alpha.$$

The lexicon of the language consists of the following three sets of nonlogical symbols: TC, a set of time point symbols; FACTS, a set of fact symbols; EVENTS, a set of event symbols.

The set of of well-formed formulas combining the logical and nonlogical symbols is recursively defined as follows:

1. If $t_1, t_2 \in TC$ then $t_1 = t_2$, $t_1 \preceq t_2$, and $t_1 \prec t_2$ are wffs.
2. If $t_1, t_2 \in TC$ and $fa \in FACTS$ then $HOLDS(t_1, t_2, fa)$ is a wff.
3. If $t_1, t_2 \in TC$ and $ev \in EVENTS$ then $OCC(t_1, t_2, ev)$ is a wff.
4. If $\phi_1$ and $\phi_2$ are wffs then so are $\phi_1 \wedge \phi_2$, $\phi_1 \vee \phi_2$, and $\neg\phi_1$.
5. If $\phi$ is a wff and $t \in TC$ then $\Box_t(\phi)$ is a wff.
6. If $\phi_1, ..., \phi_k$ are wffs, $\theta_0, ..., \theta_k$ are real numbers, and $t \in TC$ then $\theta_0 P_t(\phi_1) \cdot ... \cdot P_t(\phi_k) + \theta_1 P_t(\phi_1) \cdot ... \cdot P_t(\phi_{k-1}) + ... + \theta_k \geq 0$ is a wff.

The four example sentences from the introduction can be represented as follows. In the first three sentence below $t_0$, $t_1$, $t_2$, $t_3$ are time points such that $(t_0 \prec t_1 \prec t_2 \prec t_3)$ and $t_0$ denotes the time "now".

- It is not possible for me to raise and lower my arm at the same time.
  $\neg\Diamond_{t_0}[OCC(\text{raise}, t_1, t_2) \wedge OCC(\text{lower}, t_1, t_2)]$
- If the oven temperature is constant while I am baking the souffle, the souffle is likely to turn out right.

$$P_{t_0}(HOLDS(\text{souffle-done-right}, t_2, t_2) | OCC(\text{bake-souffle}, t_1, t_2) \wedge HOLDS(\text{oven-temp-const}, t_1, t_2)) \geq .9$$

- There is a 50% chance that at noon the train crash will be inevitable.
  $(t_1 = \text{noon}) \wedge P_{t_0}(\Box_{t_1} OCC(t_2, t_3, \text{crash})) = .5$

In the next sentence $t_0$, $t_1$, and $t_2$ are all time points in the past and $(t_0 \prec t_1 \prec t_2)$.

- I won the lottery even though it was unlikely.

$$P_{t_0}(OCC(t_1, t_2, \text{win-lottery})) = .0001 \wedge OCC(t_1, t_2, \text{win-lottery})$$

## 4 Semantics

A model is a 7-tuple $\langle W, T, FA, EV, R, PR, F\rangle$, where:

- $W$ is the set of possible world-histories.
- $T$ is a totally ordered set of time points, corresponding to the reals.
- $FA$ and $EV$ range over subsets of $2^{(T\times T)\times W}$ designating the sets of domain elements of type fact and event, respectively.
- $R$ is a relation defined on $T \times W \times W$. $R(t, w_1, w_2)$ means that world-histories $w_1$ and $w_2$ share a common past up to time $t$. The set of all world-histories accessible from $w$ at time $t$ will be designated $R_t^w$.
- $PR$ is a probability assignment function that assigns to each time $t$ and world-history $w$ a probability function $\mu_t^w$.

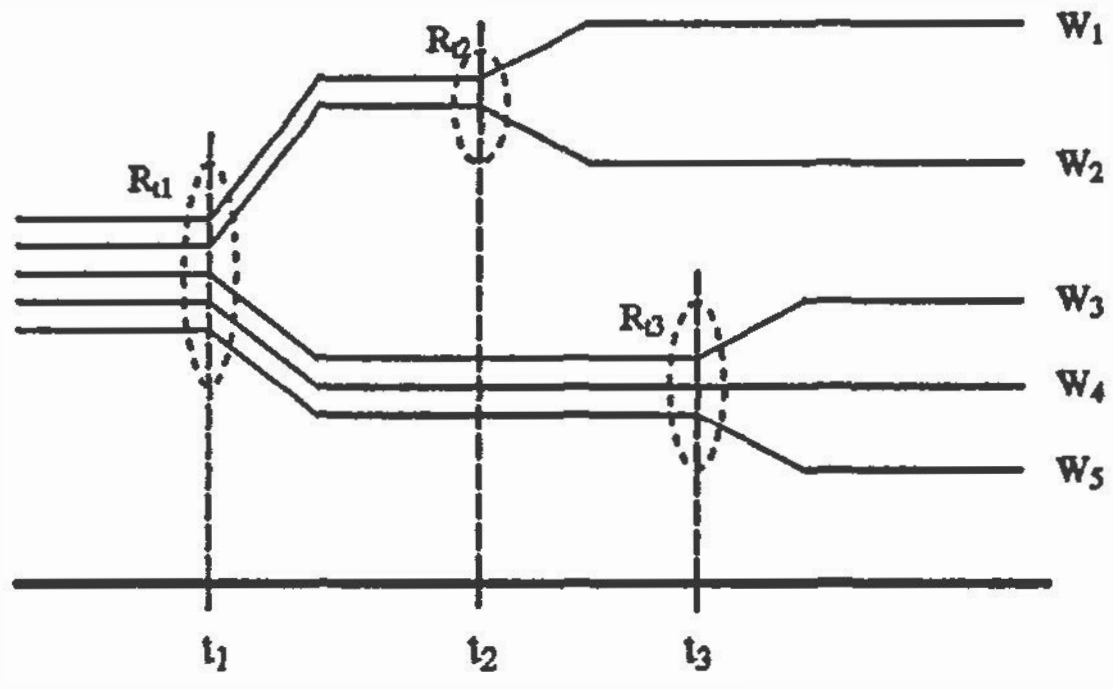

Figure 1: Structure imposed by the R accessibility relation.

- $F$ is the denotation function that maps every time symbol to an element of $T$, every fact symbol to an element of $FA$, and every event symbol to an element of $EV$.

In section 2 the ontology of the logic was discussed from an intuitive standpoint. In order to obtain the desired intuitive properties, a number of constraints must be imposed on the models. These constraints C1-6 are presented in the following discussion.

The future branching temporal tree is defined in terms of the $R$ relation over world-histories. $R$ relates worlds-histories with common pasts. To capture the future branching nature of time we say that if two world-histories share a common past up to time $t_2$ then they share a common past up to any earlier time:

(C1) If $t_1 \leq t_2$ and $R(t_2, w_1, w_2)$ then $R(t_1, w_1, w_2)$.

Since $R$ just represents the equality of histories up to a time $t$, for a fixed time $R$ is an equivalence relation.

(C2) For a fixed time $R$ is reflexive, transitive, and symmetric.

Figure 1 illustrates how the $R$ relation ties together the different world-histories to form the temporal tree structure.

Facts and events hold and occur in various world-histories at various times. Thus, we identify facts and events with sets of ⟨temporal interval, world⟩ pairs. Following Shoham [1987], we take a temporal interval to be simply an ordered pair of time points. So a fact(event) is a set of elements of the form $\langle\langle t_1, t_2\rangle, w\rangle$, where $t_1, t_2 \in T$, $t_1 \leq t_2$, and $w \in W$. If $\langle\langle t_1, t_2\rangle, w\rangle \in A$ then fact(event) $A$ holds(occurs) during interval $\langle t_1, t_2\rangle$ in world-history $w$.

As mentioned earlier, facts and events differ in their temporal properties. This distinction is captured by the following semantic constraint which states that if a fact holds over an interval, it holds over all subintervals:

(C3) If $t_1 \leq t_2 \leq t_3 \leq t_4$, $t_1 \neq t_3$, $t_2 \neq t_4$, $fa \in FA$ and $\langle\langle t_1, t_4\rangle, w\rangle \in fa$ then $\langle\langle t_2, t_3\rangle, w\rangle \in fa$.

To ensure consistency, the $R$ accessibility relation must be compatible with the specifications in the model describing the facts that hold at different times and events that occur at different times in each world-history. Because $R$ relates world-histories with common pasts, if two world-histories are $R$ related at time $t$, they must agree on all facts(events) that hold(occur) over intervals ending before or at the same time as $t$:

(C4) If $t_1 \leq t_2$ and $R(t_2, w_1, w_2)$ then $\langle\langle t_0, t_1\rangle, w_1\rangle \in A$ iff $\langle\langle t_0, t_1\rangle, w_2\rangle \in A$.

In section 2, two desired characteristics of the probability operator were mentioned. The first is that the probability of a past fact or event should be either zero or one, depending on whether or not it actually happened. This is achieved by the following constraint.

(C5) $\mu_t^w(R_t^w) = 1$.

Defining the probabilities in this way makes good intuitive sense if we look at the meaning of $R$. $R_t^w$ designates a set of world-histories that are objectively possible with respect to $w$ at time $t$. It is natural that the set of world-histories that are objectively likely with respect to $w$ at time $t$ should be a subset of the ones that are possible.

The second desired characteristic is that the probability at a time $t$ be completely determined by the history up to that time. In other words, worlds that share a common past up to a given time $t$ should share a common probability function at that time. This is again easily captured in terms of the $R$ accessibility relation:

(C6) If $w_1, w_2 \in R_t^w$ then $\mu_t^{w_1} = \mu_t^{w_2}$.

Given the models described above, the semantic definitions for the well-formed formulas are defined as follows. Note that the denotation of an expression $\phi$ relative to a model $M$ and a world-history $w$ is designated by $[\![\phi]\!]^{M,w}$.

1. $[\![t_1 \preceq t_2]\!]^{M,w}$ =true iff $F(t_1) \leq F(t_2)$.
2. $[\![t_1 = t_2]\!]^{M,w}$ = true iff $F(t_1) = F(t_2)$.
3. $[\![HOLDS(t_1, t_2, fa)]\!]^{M,w}$ = true iff $\langle\langle F(t_1), F(t_2)\rangle, w\rangle \in F(fa)$.
4. $[\![OCC(t_1, t_2, ev)]\!]^{M,w}$ = true iff $\langle\langle F(t_1), F(t_2)\rangle, w\rangle \in F(ev)$.
5. $[\![\Box_t(\phi)]\!]^{M,w}$ = true iff $[\![\phi]\!]^{M,w'}$ = true for every $w'$ such that $R(F(t), w, w')$.
6. $[\![P_t(\phi) \geq \alpha]\!]^{M,w}$ = true iff $\mu_t^w(\{w' \in R_t^w \mid [\![\phi]\!]^{M,w'} = \text{true}\}) \geq \alpha$.

The interesting definitions are the last two. Definition 5 says that a sentence is inevitable in a world $w$ at a time $t$ iff it is true in all worlds that share a common history with $w$ up to time $t$. Definition 6 says that the probability of a sentence $\phi$ is at least $\alpha$ in a world $w$ at a time $t$ iff the probability of those accessible worlds in which $\phi$ is true is at least $\alpha$.

A sentence $\phi$ is *satisfied* by a model $M$ at a world $w$ if it is assigned the value true in that model and world. A sentence is *valid* if it is satisfied by every model at every world.

The following are some examples of valid sentences.

- The past is determined:

$$(t_1 \preceq t_2) \rightarrow [P_{t_2}(HOLDS(t_0,t_1,\phi)) = 0 \vee P_{t_2}(HOLDS(t_0,t_1,\phi)) = 1]$$

$$(t_1 \preceq t_2) \rightarrow [P_{t_2}(OCC(t_0,t_1,\phi)) = 0 \vee P_{t_2}(OCC(t_0,t_1,\phi)) = 1]$$

- If something is inevitable, it is certain:

$$\Box_t(\phi) \rightarrow P_t(\phi) = 1$$

- Inevitability persists:

$$(t_1 \preceq t_2) \rightarrow [\Box_{t_1}(\phi) = 1 \rightarrow \Box_{t_2}(\phi) = 1]$$

- We also have the following rule of probabilistic detachment:

$$\text{If } \phi \models \psi \text{ then } P_t(\psi) \geq P_t(\phi)$$

## 5 Miller's Principle

As a consequence of the characteristics we have imposed on probability, there is an interesting relationship between probabilities assigned to the same sentence at various times. The relationship is that the probability of a sentence at some time given that its probability at some future time is at least $\alpha$ should be at least $\alpha$: $P_{t_1}(\phi \mid P_{t_2}(\phi) \geq \alpha) \geq \alpha$. This relation is called *Miller's principle* and several nontemporal variants of it were first suggested by Brian Skyrms [1980] as possible constraints on higher-order probabilities. We give here an informal intuitive argument that Miller's principle holds in this logic. The formal proof is omitted due to space limitations.

If we agree that the history of events determines chance, as discussed in section 2, then temporally qualified probabilities can be written as conditional probabilities: $\mu_t(A) = \mu(A|\text{history up to } t)$. Then what should be the chance of $A$ at time $t_1$ given that events unfold to time $t_2$, $t_1 \leq t_2$, such that the chance of $A$ at $t_2$ is $\alpha$? At time $t_1$, as far as events that determine the chance of A are concerned, all possible futures up to time $t_2$ are equivalent. So the probability should just be $\alpha$:

$$\begin{aligned}
&\mu_{t_1}(A \mid \mu_{t_2}(A) = \alpha) = \\
&\mu_{t_1}(A \mid \mu(A \mid \text{history up to } t_2) = \alpha) = \\
&\mu_{t_1}(A \mid \text{hist up to } t_2 \text{ is such that } \mu(A) = \alpha) = \\
&\mu(A \mid \text{history up to } t_1\ \& \\
&\quad(\text{history up to } t_2 \text{ is such that } \mu(A) = \alpha)) = \\
&\mu(A \mid \text{hist up to } t_2 \text{ is such that } \mu(A) = \alpha) = \alpha
\end{aligned}$$

In the object language, Miller's principle can be stated as the sentence schema:

$$(t \preceq t') \rightarrow P_t(\phi \wedge P_{t'}(\phi) \geq \alpha) \geq \alpha \cdot P_t(P_{t'}(\phi) \geq \alpha)$$

Every instance of this schema is valid in the temporal probability models.

As a direct consequence of Miller's principle, the probability at a given time is the expected value of the probability at any future time. This will be called the *expected value principle*. Miller's principle and the expected value principle are useful because they allow the current probabilities of facts and events to be inferred from the probabilities of their future probabilities. For example, suppose that I am first going to randomly choose between two coins, one fair coin and one with a 70% chance of heads, and then I am going to flip the chosen coin. There is a 50% chance that the coin will have a 70% chance of landing heads and a 50% chance that the coin will have a 50% chance of landing heads. By the expected value principle, it follows that there is now a 60% chance that the coin flip will result in heads.

## 6 Causality

It was stated in the introduction that one of the intended uses of this logic is for reasoning about action effects. The primary relationship between an action and its effects is the causal relationship. Suppes [1970] provides an elegant theory of probabilistic causality. He gives three "prima facie" conditions which are necessary for event A to cause event E. Note that his theory was developed in the context of instantaneous events. In the sentences below, $A_{t'}$ denotes that event $A$ occurs at time $t'$ and $E_t$ denotes that event $E$ occurs at time $t$.

1. Temporal precedence: $t' \prec t$
2. Possibility of cause: $P(A_t) > 0$
3. Positive influence: $P(E_t|A_{t'}) > P(E_t)$

These three conditions can be restated in the logic of time intervals. The three conditions stated for event $A$ that occurs in the interval $t_A$ to $t_{A'}$ and event $E$ that occurs in the interval $t_E$ to $t_{E'}$ are

1. Temporal non-succession: $t_A \prec t_{E'}$
2. Possibility of cause: $P_{t_A}(OCC(A,t_A,t_{A'})) > 0$
3. Positive influence:

$$P_{t_A}(OCC(E,t_E,t_{E'})|OCC(A,t_A,t_{A'})) > P_{t_A}(OCC(E,t_E,t_{E'}))$$

Now it can be shown that because of the way objective probability has been defined, condition 3) entails both conditions 1) and 2). First, if condition 3) is expanded out into its proper form in the logic it becomes.

$$P_{t_A}(OCC(E, t_E, t_{E'}) \wedge OCC(A, t_A, t_{A'})) > P_{t_A}(OCC(E, t_E, t_{E'})) \cdot P_{t_A}(OCC(A, t_A, t_{A'}))$$

It is clear that if $P_{t_A}(OCC(A, t_A, t_{A'})) = 0$ then the sentence is false. So if 3) holds, 2) must hold. Next, if 1) is false then $t_{E'} \preceq t_A$. Since the probability of past events is either zero or one, either

$$P_{t_A}(OCC(E, t_E, t_{E'})) = 0$$

or

$$P_{t_A}(OCC(E, t_E, t_{E'})) = 1.$$

Either case contradicts 3). So if 3) holds, 1) must also hold.

This result shows that the model of objective probability has captured the temporal flow of causality as intended - actions cannot affect the past. As a consequence of this result, if we use condition 3) to define what it means for a plan to achieve a goal then we can prove that actions after the time of the goal cannot contribute to achieving the goal.

The ability of the logic to represent and distinguish between truth and probability allows us to distinguish between potential causation and actual causation. Suppes [1970, pages 37-41] defines actual causation as potential causation that actually occurs:

$$OCC(E, t_E, t_{E'}) \wedge P_{t_A}(OCC(E, t_E, t_{E'}) | OCC(A, t_A, t_{A'})) > P_{t_A}(OCC(E, t_E, t_{E'}))$$

## 7 Examples

In this section two examples illustrating the use of the logic are presented. The first example involves reasoning about the state of the world and the influence of the world state on the consequences of actions. Suppose I own a car that is not very reliable and often does not start if the weather is too cold. My car starts 30% of the time when the temperature is below freezing. Suppose that there is an 80% chance the temperature will be below freezing tomorrow morning. What is the chance that my car will start tomorrow morning? The first sentence can be represented as:

$$P_{t_0}(OCC(t_s, t_{s'}, \text{start}) \mid OCC(\text{turn-key}, t_s, t_{s'}) \wedge HOLDS(t_s, t_{s'}, \text{below-freezing})) = .3$$

The second sentence can be represented as:

$$P_{t_0}(HOLDS(t_M, t_{M'}, \text{below-freezing})) = .8$$

We also know that I will try to start my car sometime during tomorrow morning:

$$(t_0 \prec t_M \preceq t_s \prec t_{s'} \preceq t_{M'})$$

By the property of facts holding over subintervals and the rule of probabilistic detachment,

$$P_{t_0}(HOLDS(t_s, t_{s'}, \text{below-freezing})) \geq .8$$

Assuming that the action of turning the key and the temperature are independent, we have

$$P_{t_0}(OCC(t_s, t_{s'}, \text{start}) \wedge OCC(\text{turn-key}, t_s, t_{s'})) = (.3)(.8) = .24$$

Finally, Because the probability that my car will start given that the temperature is not below freezing is at most one, we have by the law of total probability

$$P_{t_0}(OCC(t_s, t_{s'}, \text{start})) \leq .24 + (1)(.2) = .44$$

Thus we can conclude that, more likely than not, my car will not start and I should probably think about an alternative mode of transportation.

The following example is a modified version of an example presented by Pelavin [1988]. It illustrates how the logic can be used to reason about the feasibility of plans. Suppose that I am going to fly to San Francisco this evening. I have two small bags and would like to carry them both onto the plane with me. In most cases it is not possible to take two carry-on bags if the plane is full. There is a 50% chance that the plane will be full this evening. What is the chance that carrying both bags on simultaneously will be a feasible course of action? The situation can be described by the following three sentences

$$P_{now}(\neg\Diamond_{t_1}[OCC(\text{carry-b1}, t_1, t_2) \wedge OCC(\text{carry-b2}, t_1, t_2)] \mid HOLDS(\text{plane-full}, t_1, t_2)) = .8$$

$$P_{now}(HOLDS(\text{plane-full}, t_1, t_2)) = .5$$

$$now \prec t_1 \prec t_2$$

One possible model satisfying these sentences is shown in figure 2. The labels "OCCURS" and "¬OCCURS" designate the co-occurrence and non co-occurrence of the two actions, respectively. Note that in worlds $w_1$ - $w_4$ we have

$$\neg\Diamond_{t_1}[OCC(\text{carry-b1}, t_1, t_2) \wedge OCC(\text{carry-b2}, t_1, t_2)]$$

and in $w_5$ and $w_6$ we have

$$\Diamond_{t_1}[OCC(\text{carry-b1}, t_1, t_2) \wedge OCC(\text{carry-b2}, t_1, t_2)].$$

From the first two sentences it follows that

$$P_{now}(\neg\Diamond_{t_1}[OCC(\text{carry-b1}, t_1, t_2) \wedge OCC(\text{carry-b2}, t_1, t_2)]) \geq .4$$

So there is at least a 40% chance that carrying both bags simultaneously will not be feasible.

Furthermore, an upper bound on the current probability of the co-occurence of the two actions can be

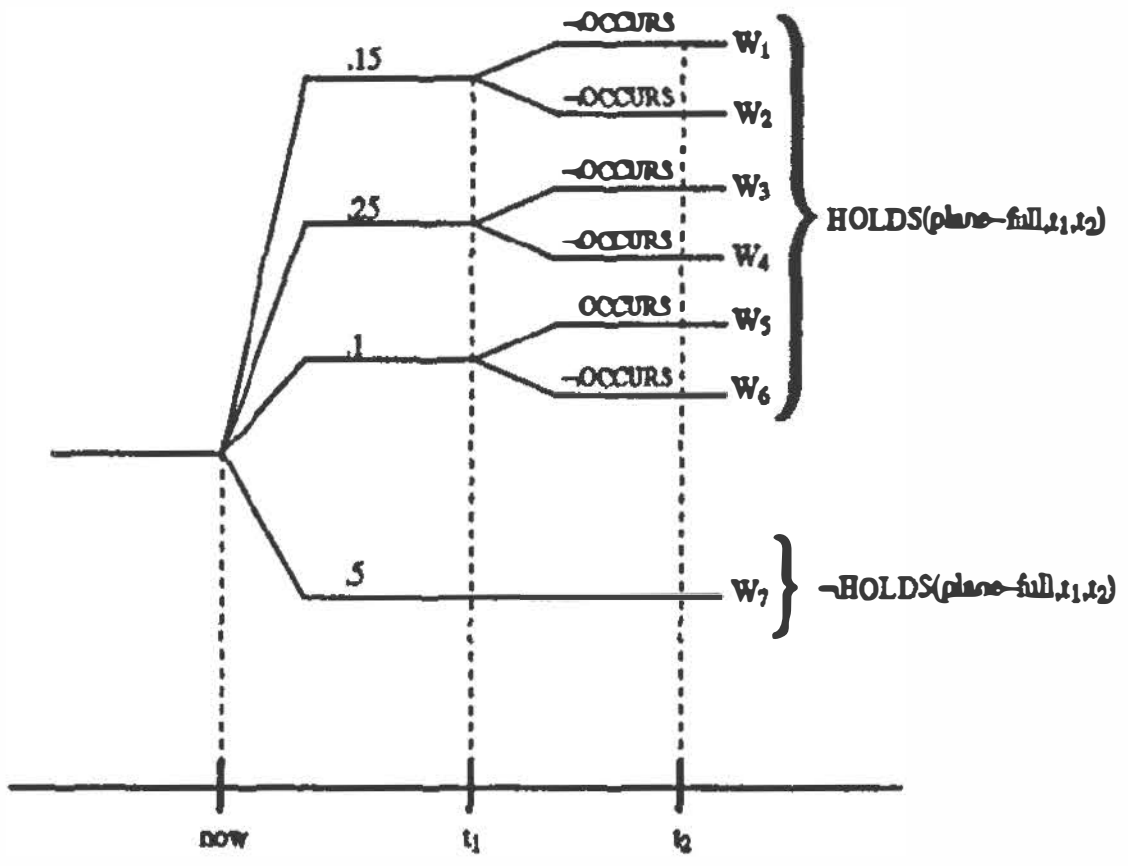

Figure 2: Possible model for carry example.

calculated. By the relation between possibility and probability

$$P_{now}(P_{t_1}(OCC(\text{carry-b1}, t_1, t_2) \wedge OCC(\text{carry-b2}, t_1, t_2)) = 0) \geq .4$$

and by the expected value property

$$P_{now}(OCC(\text{carry-b1}, t_1, t_2) \wedge OCC(\text{carry-b2}, t_1, t_2)) \leq .6$$

## 8 Related Work

The logic presented here is closely related to Pelavin's [1988] temporal planning logic. Pelavin develops a future-branching time logic for reasoning about planning problems involving concurrent actions and external events. He starts with Allen's [1984] linear temporal logic of time intervals and extends it with two modal operators, *INEV* and *IFTRIED*, to reason about future branching time and action effects, respectively. *INEV* is exactly our □ operator. *IFTRIED* is a counterfactual operator that associates the attempt of an action with the truth of a sentence. The semantics of the operator are based on Stalnaker's and Lewis's theories of counterfactuals. *IFTRIED* captures the temporal relation of action and effect—an action cannot affect the state of the world at any time preceding its attempt. Effects of general events cannot be represented in the logic.

The present work departs from Pelavin's framework in two major ways. First, points are taken to be the primitive temporal objects rather than intervals. Pelavin[1988, p84] himself notes that this results in a more natural definition of the accessibility relation. Second, and more importantly, the language can represent uncertainty. Representing uncertainty with objective probability eliminates the need for a separate counterfactual operator and its semantic counterpart: the similarity measure over worlds. The standard deterministic counterfactual operator is replaced by conditional probability. Skyrms [1980] provides an elegant probabilistic account of counterfactuals based on the notion of objective probability.

The temporal models we have presented are essentially those of van Fraassen[1980]. He presents a semantic theory that models subjective probability and objective chance, using a future-branching model of time points. In his models, objective probability can change with time but truth values cannot. He shows that a property equivalent to Miller's principle holds between subjective probability and objective chance but not between objective chance at different times. He also does not provide a logical language. The probability logic we have presented is patterned after Fagin, Halpern, and Megiddo's [1988] logic. They discuss axiomatizations and decision procedures for various probability logics. The logics presented are not formulated in a temporal framework.

## 9 Future Research

This paper has presented a logic for reasoning about objective probability. It is important for a planning representation to be able to represent the state of knowledge of the planning agent. This can be done by introducing subjective probabilities into the logic. Several philosophers have discussed the problem of combining subjective probability and objective chance [Skyrms, 1980, Appendix 2][Lewis, 1980; van Fraassen, 1980]. The general consensus is that agents have subjective beliefs concerning objective chance and the two are related by certain constraints, although there is disagreement as to precisely what the constraints should be. It can be shown that such a hybrid representation of beliefs is necessary in order to make rational decisions in some cases where causality is a factor [Skyrms, 1980, ch IIC][Lewis, 1981; Maher, 1987]. Extending the logic with subjective probability is relatively straightforward. Subjective probability can be modeled by defining probability functions over all worlds, not just the accessible ones. Corresponding to the two types of probability in the models, there would be two probability operators in the object language.

Previous work [Halpern, 1989; Haddawy and Frisch, 1990] has shown that probability logic can be viewed as a generalization of modal logic. It has been shown [Gaifman, 1988; Haddawy and Frisch, 1990] that the logics corresponding to staged probability models, similar to the temporal probability models presented here, are closely related to certain modal logics. It would be interesting to see whether the probability logic presented in this paper corresponds exactly to some temporal modal logic. Such a modal logic would be useful in providing a quali-tative representation of probabilistic information.

The logic presented in this paper is propositional. Adding a theory of quantification would greatly enhance the expressive power of the language.

In this paper no distinction has been made between the representation of actions and other events. But there are important distinctions. Actions have events associated with them but additionally actions are performed by an agent. An agent attempts an action and if the conditions are right, the action occurs. For example, I attempt to lift an object and if the object is not too heavy, I succeed in lifting it. The author is currently elaborating the ontology presented in this paper to distinguish between action attempts and action occurrences.